\documentclass[sigconf]{acmart}
\acmSubmissionID{1024}

\usepackage{booktabs} 
\usepackage{float}
\usepackage{multirow}
\usepackage{bm}

\usepackage[ruled]{algorithm2e} 

\SetAlFnt{\small}
\SetAlCapFnt{\small}
\SetAlCapNameFnt{\small}
\SetAlCapHSkip{0pt}

\makeatletter
\let\@authorsaddresses\@empty
\makeatother

\copyrightyear{2024}
\acmYear{2024}
\setcopyright{rightsretained}
\acmConference[SIGGRAPH Conference Papers '24]{Special Interest Group on Computer Graphics and Interactive Techniques Conference Conference Papers '24}{July 27-August 1, 2024}{Denver, CO, USA}
\acmBooktitle{Special Interest Group on Computer Graphics and Interactive Techniques Conference Conference Papers '24 (SIGGRAPH Conference Papers '24), July 27-August 1, 2024, Denver, CO, USA}
\acmDOI{10.1145/3641519.3657507}
\acmISBN{979-8-4007-0525-0/24/07}

\begin{document}
\title{Deep Hybrid Camera Deblurring for Smartphone Cameras}



\author{Jaesung Rim}
\orcid{0009-0001-3101-390X}
\affiliation{%
  \institution{POSTECH}
   \country{South Korea}
}
\email{jsrim123@postech.ac.kr}

\author{Junyong Lee}
\authornote{Work done prior to joining Samsung AI Center Toronto.}
\orcid{0000-0001-6472-0582}
\affiliation{%
  \institution{Samsung AI Center Toronto}
   \country{Canada}
}
\email{j.lee8@samsung.com}

\author{Heemin Yang}
\orcid{0009-0004-2891-6093}
\affiliation{%
  \institution{POSTECH}
   \country{South Korea}
}
\email{heeminid@postech.ac.kr}

\author{Sunghyun Cho}
\orcid{0000-0001-7627-3513}
\affiliation{%
  \institution{POSTECH}
   \country{South Korea}
}
\email{s.cho@postech.ac.kr}

\renewcommand\shortauthors{Rim et al}
\newcommand{\Eq}[1]  {Eq.\ (\ref{eq:#1})}
\newcommand{\Eqs}[1] {Eqs.\ (\ref{eq:#1})}
\newcommand{\Fig}[1] {Fig. \ref{fig:#1}}
\newcommand{\Figs}[1] {Figs. \ref{fig:#1}}
\newcommand{\Tbl}[1]  {Table \ref{tab:#1}}
\newcommand{\Tbls}[1] {Tables \ref{tab:#1}}
\newcommand{\Sec}[1] {Sec.\ \ref{sec:#1}}
\newcommand{\Secs}[1] {Secs.\ \ref{sec:#1}}
\newcommand{\etal}   {{et al.}}
\newcommand{\Etal}   {{et al.}}

\definecolor{brown}{rgb}{0.65, 0.16, 0.16}
\definecolor{purp}{rgb}{0.65, 0.16, 0.65}
\definecolor{orange}{rgb}{1.0, 0.5, 0.0}
\definecolor{blue}{rgb}{0.0, 0.5, 1.0}
\definecolor{green}{rgb}{0, 0.7, 0.2}
\definecolor{lgreen}{rgb}{0.6, 0.8, 0}
\definecolor{red}{rgb}{1.0, 0, 0}
\definecolor{darkblue}{rgb}{0, 0.2, 0.6}

\newcommand{\W} {$W$}
\newcommand{\UW} {$\mathbf{U}$}

\newcommand{\son}[1]{{\textcolor{magenta}{lee: #1}}}
\newcommand{\sean}[2]{{\textcolor{green}{sean: #1}\textcolor{magenta}{#2}}}
\newcommand{\sunghyun}[1]{{\textcolor[rgb]{0.6,0.0,0.6}{sunghyun: #1}}}
\newcommand{\heemin}[1]{{\textcolor{cyan}{heemin: #1}}}
\newcommand{\rjs}[1]{{\textcolor[rgb]{1,0.0,0.0}{#1}}}
\newcommand{\rjsc}[1]{{\textcolor[rgb]{0.6,0.6,1}{#1}}}
\newcommand{\kyoungkook}[1]{{\textcolor[rgb]{0.6,0.6,1}{kyoungkook: #1}}}
\newcommand{\jy}[1]{{\textcolor[rgb]{0.2,0.2,1}{Junyong: #1}}}
\newcommand{\jytodo}[1]{{\color{red}#1}}
\newcommand{\change}[1]{{\color{red}#1}}


\newcommand{\argmin}{\mathop{\mathrm{argmin}}\limits} 


\def\ie{i.e.}
\def\eg{e.g.}
\def\wrt{\emph{w.r.t.}}

\begin{teaserfigure}
    \centering
    \includegraphics[width=1.0\textwidth]{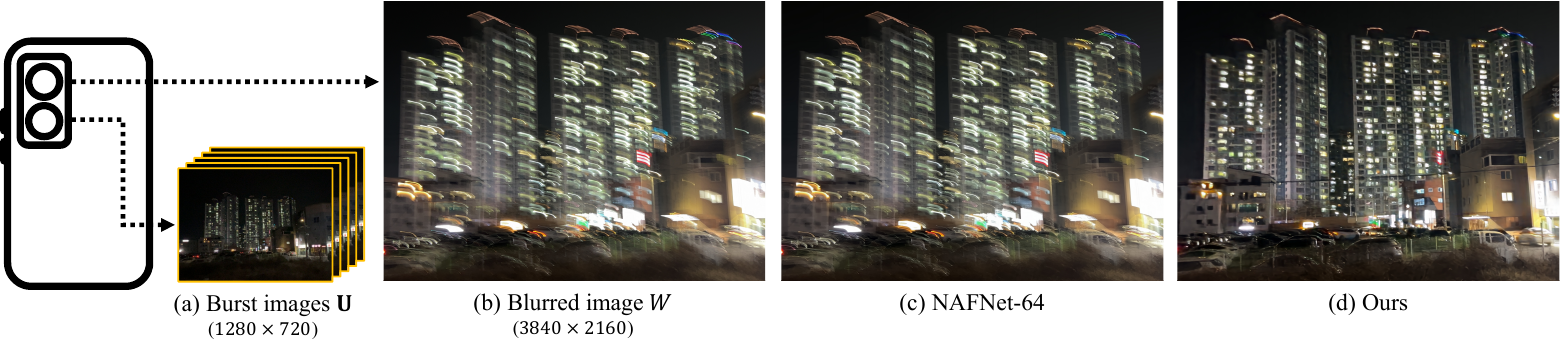}
    \setlength{\abovecaptionskip}{-2.0mm}
    \caption{Our hybrid camera system and deblurred results. We simultaneously capture a long-exposure wide image $W$ and short-exposure burst ultra-wide images $\mathbf{U}$ from a smartphone, and utilize the burst images to deblur $W$. Our method produces significantly sharper results compared to NAFNet-64~\cite{chen2022simple}, a state-of-the-art single-image deblurring method.}
    \label{fig:teaser}
\end{teaserfigure}

\begin{abstract}
Mobile cameras, despite their significant advancements, still have difficulty in low-light imaging due to compact sensors and lenses, leading to longer exposures and motion blur. 
Traditional blind deconvolution methods and learning-based deblurring methods can be potential solutions to remove blur. However, achieving practical performance still remains a challenge. 
To address this, we propose a learning-based deblurring framework for smartphones, utilizing wide and ultra-wide cameras as a hybrid camera system. 
We simultaneously capture a long-exposure wide image and short-exposure burst ultra-wide images, and utilize the burst images to deblur the wide image.
To fully exploit burst ultra-wide images, we present HCDeblur, a practical deblurring framework that includes novel deblurring networks, HC-DNet and HC-FNet.
HC-DNet utilizes motion information extracted from burst images to deblur a wide image, and HC-FNet leverages burst images as reference images to further enhance a deblurred output.
For training and evaluating the proposed method, we introduce the HCBlur dataset, which consists of synthetic and real-world datasets.
Our experiments demonstrate that HCDeblur achieves state-of-the-art deblurring quality. Code and datasets are available at \color{blue}{\url{https://cg.postech.ac.kr/research/HCDeblur}}.
\end{abstract}

%
%

\ccsdesc[500]{Computing methodologies~Computational photography}

%
%

\keywords{motion deblurring, hybrid camera fusion, mobile imaging, deep neural networks}

\maketitle

\section{Introduction}
\label{sec:intro}

While mobile cameras have significantly improved, they still struggle in low-light environments due to their small sensors and lenses, leading to longer exposure time and motion blur from hand movement or moving objects.
To remove blur, singe-image deblurring methods~\cite{Fergus-SIGGRAPH06,Shan-SIGGRAPH08,Cho-SIGGRAPHAsia09,Xu-ECCV10,Xu-CVPR13,Pan-CVPR16,Sun-ICCP13} have been widely studied.
Recently, learning-based deblurring methods~\cite{Nah_2017_CVPR,Tao-CVPR18,DeblurGAN,DeblurGAN-v2,Zamir_2021_CVPR,Wang_2022_CVPR,Zamir_2022_CVPR,cho2021rethinking,Chen_2021_CVPR,chen2022simple} have significantly improved the deblurring performance.
Nonetheless, as shown in \Fig{teaser}-(c), the performance of single-image deblurring is still limited, especially for large blurs.

Several methods have been proposed to improve the deblurring performance by exploiting additional inputs, such as a short-exposure image and event data. 
Reference-based deblurring methods~\cite{Mustaniemi_2020_BMVC, chang2021low, zhao2022d2hnet, lai2022face} employ an extra short-exposure image as a reference image to deblur a long-exposure image. 
These methods align and fuse a blurred image and a reference image to restore a high-quality deblurred image.
However, they often struggle to restore sharp details, especially for severely blurred images, due to the challenges of aligning a severely blurred image with a reference image accurately.
Event-guided deblurring methods~\cite{cho2023non,jiang2020learning,haoyu2020learning,xu2021motion,sun2022event,kim2022event,Zhang_2023_ICCV} simultaneously capture a blurred image and event data using a specially designed dual-camera comprising an RGB camera and an event camera, and achieve significant deblurring performance utilizing motion information extracted from event data.
However, these methods require an additional event camera, which is not commonly found on most commodity cameras.

In this paper, we propose HCDeblur, a practical image deblurring framework designed for modern smartphones like the Apple iPhone and Samsung Galaxy series, which now commonly feature multi-camera systems.
Our approach utilizes such multi-camera systems as a \emph{hybrid camera system}, which consists of a primary camera and a secondary camera with a higher frame rate.
The hybrid camera system simultaneously captures a long-exposure image and a burst of short-exposure images.
These burst images provide crucial information on pixel-wise camera and object motions during the exposure time, which is challenging to acquire from a single image, while it can significantly enhance the deblurring performance.
Additionally, burst images offer high-frequency details that can complement high-frequency details lost in a blurred image.
HCDeblur leverages both motion and detail information from burst images and achieves unparalleled performance in image deblurring (\Fig{teaser}-(d)).

The concept of hybrid camera deblurring was first proposed by Ben-Ezra and Nayar~\cite{Ben-Ezra} to resolve the ill-posedness of image deblurring.
Specifically, they propose to use a low-resolution high-speed camera as a secondary camera. 
Using a low-resolution high-speed camera, they capture a burst of short-exposure images.
These images are then used to estimate a uniform motion blur kernel, which is subsequently used for the non-blind deconvolution of a blurred image taken by a primary camera.
Tai~et al.~\shortcite{Tai_hybrid,TAI_TPAMI} further extended the idea to handle non-uniform deblurring and video deblurring.
However, these approaches rely on classical blur models and optimization methods, which severely limits their performance.
In contrast, our approach is a learning-based approach that adopts a hybrid camera system. Our approach is tailored for real-world smartphone cameras and shows superior performance in real-world scenarios.

Our framework employs the wide and ultra-wide cameras of a smartphone as the primary and secondary cameras, respectively, as depicted in \Fig{teaser}. The wide camera, typically the main camera in smartphones, captures images at a slow shutter speed. Conversely, the ultra-wide camera, due to its broader field of view (FOV), can gather motion and detail information for the entire region of the image captured by the wide camera. This secondary camera simultaneously captures a burst of low-resolution images at a higher shutter speed and a high frame rate.

Once a wide image and a burst of ultra-wide images are captured, our framework estimates a deblurred image of the wide image with the aid of burst ultra-wide images.
To this end, we introduce a deep neural network equipped with two sub-networks: a Hybrid Camera Deblurring Network (HC-DNet) and a Fusion Network (HC-FNet).
HC-DNet uses burst images to construct pixel-wise blur kernels and exploits the blur kernels to obtain a deblurred image of the wide image.
While HC-DNet produces a deblurred result of superior quality compared to previous single-image deblurring methods thanks to the blur kernels, its results may still contain artifacts and remaining blur due to information loss caused by blur and inaccurate blur kernels.
To mitigate this, HC-FNet, which is inspired by burst imaging techniques~\cite{DBSR2021deep,dudhane2021burst,mehta2023gated}, refines the output of HC-DNet using the entire sequence of the burst images as reference images.

For training and evaluating our method, we also present the HCBlur dataset, which consists of two sub-datasets: HCBlur-Syn and HCBlur-Real.
HCBlur-Syn is a synthetically generated dataset for training and evaluation, and comprises 8,568 pairs of blurred wide images and ground-truth sharp images with corresponding sharp ultra-wide burst images.
On the other hand, HCBlur-Real is a real dataset for evaluation,
and provides 471 real-world pairs of a blurred wide image and burst ultra-wide images without ground-truth sharp images.
Our experimental results using the HCBlur dataset show that our method significantly outperforms state-of-the-art deblurring methods. 

To summarize, our contributions include:
\begin{itemize}
    \setlength\itemsep{0.1em}
    \item HCDebur, a learning-based hybrid camera deblurring framework specifically designed for smartphone cameras,
    \item HC-DNet and HC-FNet, which utilize burst ultra-wide images for deblurring and refining an input wide image, respectively, and
    \item the HCBlur dataset for training and evaluating hybrid camera deblurring methods. 
\end{itemize}

\section{Related Work}
\label{sec:formatting}

\paragraph{Single-image Deblurring}
Classical blind deconvolution methods \cite{Shan-SIGGRAPH08, Cho-SIGGRAPHAsia09,Xu-ECCV10,Pan-CVPR16,Cho-ICCV17,Levin-CVPR09} alternatingly estimate a blur kernel and a sharp image to deblur a blurred image.
Their performance is limited due to restrictive blur models and the ill-posedness of the deblurring problem.
Recently, learning-based methods have significantly improved the deblurring performance by learning the deblurring process from the training dataset.
Numerous deblurring networks have been proposed, such as multi-scale~\cite{Nah_2017_CVPR, Tao-CVPR18, cho2021rethinking}, multi-stage~\cite{Zhang_2019_CVPR,Zamir_2021_CVPR}, multi-scale and multi-stage~\cite{Kim2022MSSNet}, NAFNet~\cite{chen2022simple}, GAN-based~\cite{DeblurGAN-v2, DeblurGAN}, and transformer-based~\cite{Wang_2022_CVPR, tu2022maxim, Zamir_2022_CVPR} methods.
However, learning-based methods still struggle with the generalization problem due to the difficulty of learning arbitrary possible motion blur~\cite{Kaufman_2020_CVPR, Fang_2023_CVPR}.

\paragraph{Kernel-based Deblurring}  
Kernel-based deblurring methods estimate blur kernels from a blurred image and use them in deblurring networks to improve the generalization ability and performance.
Kaufman~et al.~\shortcite{Kaufman_2020_CVPR} incorporate a kernel estimation network and a deblurring network for uniform motion. 
Zhang~et al.~\shortcite{zhang2021exposure} estimate non-uniform motion offsets from a blurred image and utilize them as offsets of deformable convolution~\cite{zhu2019deformable} in a deblurring network.
Fang~et al.~\shortcite{Fang_2023_CVPR} proposed a non-uniform kernel estimation network based on the normalizing flow model and utilize estimated blur kernels for computing channel attention.
However, estimating accurate blur kernels from a single blurred image is still challenging due to the ill-posedness of the deblurring problem.

\paragraph{Reference-based Deblurring} 
Reference-based deblurring methods utilize an extra short-exposure image as a reference image.
Mustaniemi~et al.~\shortcite{Mustaniemi_2020_BMVC} proposed LSD$_2$ for jointly deblurring and denoising. 
LSFNet~\cite{chang2021low} and D2HNet~\cite{zhao2022d2hnet} employ deformable convolution~\cite{zhu2019deformable} to align and incorporate features of a blurred image and a short exposure image.
Several methods for smartphone cameras utilize extra ultra-wide images as additional inputs.
Lai~et al.~\shortcite{lai2022face} proposed a face deblurring method that leverages an ultra-wide short-exposure image to deblur a blurred wide image. 
Alzayer~et al.~\shortcite{alzayer2023dc2} proposed a defocus blur control method utilizing an extra ultra-wide image.
Our approach differs from theirs by utilizing burst reference images for deblurring, not a single reference image.

\paragraph{Hybrid Camera Deblurring} 
Ben-Ezra and Nayar~\shortcite{Ben-Ezra} first introduced the idea of hybrid camera deblurring.
Tai~et al.~\shortcite{Tai_hybrid, TAI_TPAMI} extended the idea to non-uniform deblurring and video deblurring.
Li~et al.~\shortcite{Li_hybrid} proposed a method of hybrid imaging for motion deblurring and depth map super-resolution.
Concurrent to our work, Shekarforoush~et al.~\shortcite{shekarforoush2023dualcamera} recently introduced a novel framework that also utilizes a single long-exposure image and a burst of short-exposure images to reconstruct a noise-free sharp image.
To this end, their framework integrates a kernel-based deblurring~\cite{zhang2021exposure} and a burst denoising network~\cite{DBSR2021deep}.
While their work also uses burst images as ours, ours is specifically tailored for real-world smartphone cameras, addresses misalignment between a long-exposure image and burst images, and presents realistic datasets.

\paragraph{Burst Image Enhancement} 
Burst image enhancement is a task to reconstruct a high-quality image from a burst of images.
Numerous approaches have been proposed for burst super-resolution, denoising and deblurring such as DBSR~\cite{DBSR2021deep}, BIPNet~\cite{dudhane2021burst}, Burstormer~\cite{dudhane2023burstormer}, GMTNet~\cite{mehta2023gated}, Fourier burst accumulation~\cite{Delbracio_2015_CVPR}, and a permutation-invariant network~\cite{aittala2018burst}.
While our work also employs burst images, it differs from these burst image enhancement approaches as our goal is to deblur a long-exposure wide image by utilizing extra short-exposure burst images.

\begin{figure*}[t]
    \centering      
    \includegraphics[width=0.97\linewidth]{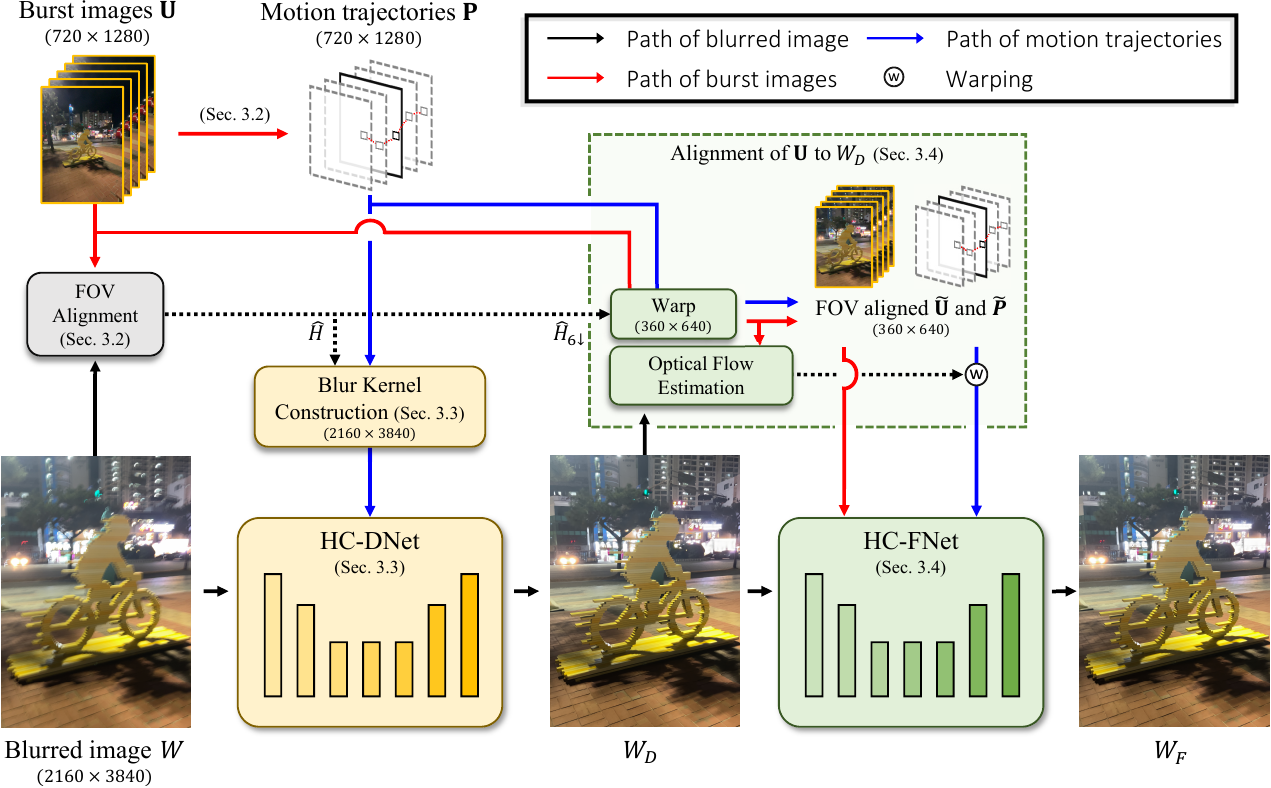}
    \setlength{\abovecaptionskip}{+1.5mm}
    \caption{
    Overview of HCDeblur. 
    Our framework takes a long-exposure wide image $W$~and a burst of short-exposure ultra-wide images $\mathbf{U}$ as inputs.  
    We estimate a homography matrix $\hat{H}$ for aligning $\mathbf{U}$ in the FOV alignment (\Sec{foval}) and compute pixel-wise motion trajectories $\mathbf{P}$ (\Sec{PT}). 
    HC-DNet performs kernel-based deblurring by exploiting blur kernels $\mathbf{K}$ constructed from $\mathbf{P}$ (\Sec{hc-dnet}). 
    After deblurring, an additional alignment step is adopted to align the burst images to the deblurred wide image $W_{D}$.
    HC-FNet further enhances the deblurred image by using the entire sequence of the burst images as reference images (\Sec{hc-fnet}). 
    }
\label{fig:pipeline}
\end{figure*}

\section{Deep Hybrid Camera Deblurring}

\Fig{pipeline} shows an overview of HCDeblur.
HCDeblur takes a long-exposure wide image $W$ and a burst of short-exposure ultra-wide images $\mathbf{U}=\{U_1, \cdots, U_N\}$ captured by our hybrid camera system (\Sec{camera_system}) as input, where $U_i$ is the $i$-th ultra-wide image, and $N$ is the number of ultra-wide images.
The wide and ultra-wide cameras have different FOVs and optical centers, resulting in geometric misalignment between them.
Thus, to obtain blur kernels from $\mathbf{U}$ that are aligned to $W$, we align $W$ and $\mathbf{U}$ using the FOV alignment (\Sec{foval}).
We also compute pixel-wise motion trajectories across $\mathbf{U}$ (\Sec{PT}) for constructing blur kernels.
Then, we perform kernel-based deblurring in HC-DNet (\Sec{hc-dnet}) using the blur kernels constructed from the motion trajectories and obtain a deblurred image $W_{D}$.
We further refine $W_{D}$ using the detail information in $\mathbf{U}$ using HC-FNet (\Sec{hc-fnet}) and obtain a final result $W_{F}$.
In the following, we describe each component of HCDeblur in more detail.

\subsection{Hybrid Camera System}
\label{sec:camera_system}

For the hybrid camera system, we employ the wide and ultra-wide cameras of an Apple iPhone 13 Pro Max in our experiments, whose focal lengths are 26mm and 13mm, respectively.
To simultaneously capture $W$ and $\mathbf{U}$, we developed an iOS app using the multi-camera API~\cite{multicamAPI}.
Our app captures a wide image at 4K resolution ($2160\times 3840$) with exposure times ranging from $1/15$ to $1/2$ sec..
At the same time, our app captures a burst of ultra-wide images of a resolution of 720P ($720\times1280$) at 60 frames per sec. (FPS). 
As wide images have a twice longer focal length, and $3\times$ more pixels along the horizontal and vertical axes than ultra-wide images,
the objects in wide images appear $6\times$ larger than those in ultra-wide images.
To prevent blur in burst images, we limit the maximum exposure time of the ultra-wide camera to $1/120$ sec..
Due to hardware limitations, the exposure times of cameras are not perfectly synchronized.
Thus, for synchronizing images from the two cameras, we also record the timestamps of the beginning and end of the exposure of both wide and ultra-wide cameras.

\subsection{FOV Alignment and Motion Estimation}
\label{sec:foval}

\paragraph{FOV Alignment}
Once a wide image $W$ and a burst of ultra-wide images $\mathbf{U}$ are captured, we first align $\mathbf{U}$ to $W$ in the FOV alignment step.
To this end, we find a single homography that best aligns $W$ and $\textbf{U}$ using the plane sweep method~\cite{collins1996space}.
Specifically, we first find the best depth $\hat{d}$ by solving:
\begin{align}
    \hat{d}=\argmin_{d \in D} MSE(W, \mathbb{W}(U_\textrm{avg}, H_d)) \label{eq:uw2w2}
\end{align}
where $D$ is a set of depth candidates, $MSE$ indicates mean-squared-error, and $\mathbb{W}$ is a warping function.
$U_\textrm{avg}$ is the average image of $\textbf{U}$.
We use $U_\textrm{avg}$ to take the blur in $W$ into account.
$H_d$ is a homography for inverse warping, which is defined as $H_d=K_u E d K_w^{-1}$, where $K_u$ and $K_w$ are intrinsic matrices of the ultra-wide and wide cameras, respectively, and $E$ is a relative extrinsic matrix. $K_w$, $K_u$ and $E$ are obtained by stereo calibration.
Once $\hat{d}$ is found, we compute its homography $H_{\hat{d}}$ for aligning $\mathbf{U}$, which we denote by $\hat{H}$.

Note that alignment between sharp images and a blurred image is a challenging task, especially when the blur is large,
and this limits the performance of previous reference-based deblurring methods.
In contrast, our approach enables more accurate alignment as we can synthetically generate $U_{avg}$ from a burst of ultra-wide reference images.
It is also important to note that our FOV alignment, which uses a single homography to align $W$ and $\mathbf{U}$, may result in some residual misalignment. However, our method is designed to be robust against such issues. This is because HC-DNet employs a homography not for directly transferring details from $\mathbf{U}$ to $W$, but for aligning blur kernels. These kernels typically vary smoothly in space, making them less sensitive to minor misalignments. Furthermore, once HC-DNet produces a deblurred image $W_{D}$, we can achieve precise alignment between $W_{D}$ and $\mathbf{U}$ by estimating the optical flow between them, and use them for HC-FNet.

\paragraph{Pixel-wise Motion Trajectories Estimation}
\label{sec:PT}
Alongside the FOV alignment, we also estimate pixel-wise motion trajectories across the ultra-wide images $\mathbf{U}$, which will be used for constructing blur kernels and for the alignment of the ultra-wide images for HC-FNet.
To this end, we estimate optical flows between adjacent images in $\mathbf{U}$.
Specifically, we denote by $c$ the index of the temporally center image in $\mathbf{U}$.
Then, we estimate the optical flow from $U_{i+1}$ to $U_{i}$ for $i<c$,
and estimate the optical flow from $U_{i-1}$ to $U_{i}$ for $i>c$.
As a result, we obtain $N-1$ optical flow maps.
From the estimated optical flow maps, we construct pixel-wise motion trajectories $\mathbf{P} = \{{P}_1, \cdots, {P}_N\}$ by accumulating the optical flow maps.
Specifically, at each pixel position, $\mathbf{P}$ contains a motion trajectory consisting of $N$ displacement vectors where the $i$-th vector is a 2D displacement vector from the center image to the $i$-th image.
For optical flow estimation, we adopt RAFT~\cite{teed2020raft} in our experiments.

\subsection{Kernel-based Deblurring using HC-DNet}
\label{sec:hc-dnet}

\begin{figure}[!t]
    \centering
    \includegraphics[width=0.93\linewidth]{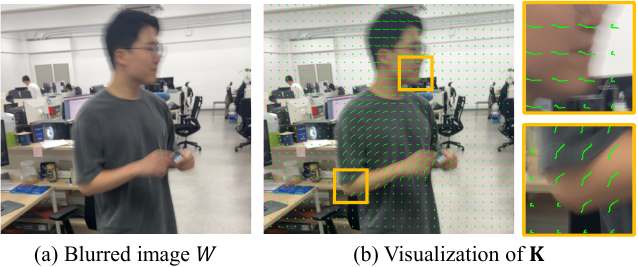}
    \setlength{\abovecaptionskip}{+1.5mm}
    \caption{Blur kernels constructed from pixel-wise motion trajectories.}
    \label{fig:blur_kernel}
\end{figure}


\paragraph{Blur Kernel Construction}
To deblur $W$ using HC-DNet, we construct blur kernels using the motion trajectories estimated from $\mathbf{U}$.
To this end, we first warp the motion trajectories using the homography $\hat{H}$ and obtain warped motion trajectories $\mathbf{\hat{P}} = \{\hat{P}_1, \cdots, \hat{P}_N\}$.
While the length of motion trajectories, $N$, varies as it is determined by the exposure time, HC-DNet takes input tensors of fixed channel sizes.
Thus, we construct blur kernels of a fixed length from the motion trajectories by resampling them.

To obtain blur kernels accurately synchronized to the exposure time of the wide image $W$, we resample the motion trajectories considering the timestamps of the wide and ultra-wide images.
Specifically, we denote the timestamps of the beginning and end of the exposure of the $i$-th ultra-wide image as $t_{i,s}$ and $t_{i,e}$.
We first compute the relative timestamp $r_i$ as:
\begin{align}
    r_i = \frac{(t_{i,s}+t_{i,e})/2 - t^{W}_{s}}{t^{W}_{s} - t^{W}_{e}}
\end{align}
where $t_s^W$ and $t_e^W$ represent the timestamps of the beginning and end of the exposure of $W$, respectively.
We then construct the blur kernel $\mathbf{K} = \{K_0, K_{0.125}, \cdots, K_{0.875}, K_1\}$ by interpolating the motion trajectories for the pre-defined set of nine timestamps.
Formally, $K_t$ is computed as:
\begin{align}
    K_{t} &= (t - r_i) \cdot \frac{\hat{P}_{i+1} - \hat{P}_{i}} {r_{i+1} - r_i} + \hat{P}_i,
\end{align}
where $t \in \{0, 0.125, \dots, 0.875, 1\}$,
$r_{i}$ and $r_{i+1}$ are the nearest relative timestamps of $\mathbf{U}$~with respect to each $t$, and
$\hat{P}_{i}$ and $\hat{P}_{i+1}$ are the displacement vector maps corresponding to $r_{i}$ and $r_{i+1}$, respectively.
Finally, we concatenate $K_t$ for all $t$ and obtain $\mathbf{K}$ of a channel size $18$.
\Fig{blur_kernel} shows an example of estimated blur kernels.

\begin{figure}[!t]
    \centering
    \includegraphics[width=1.00\linewidth]{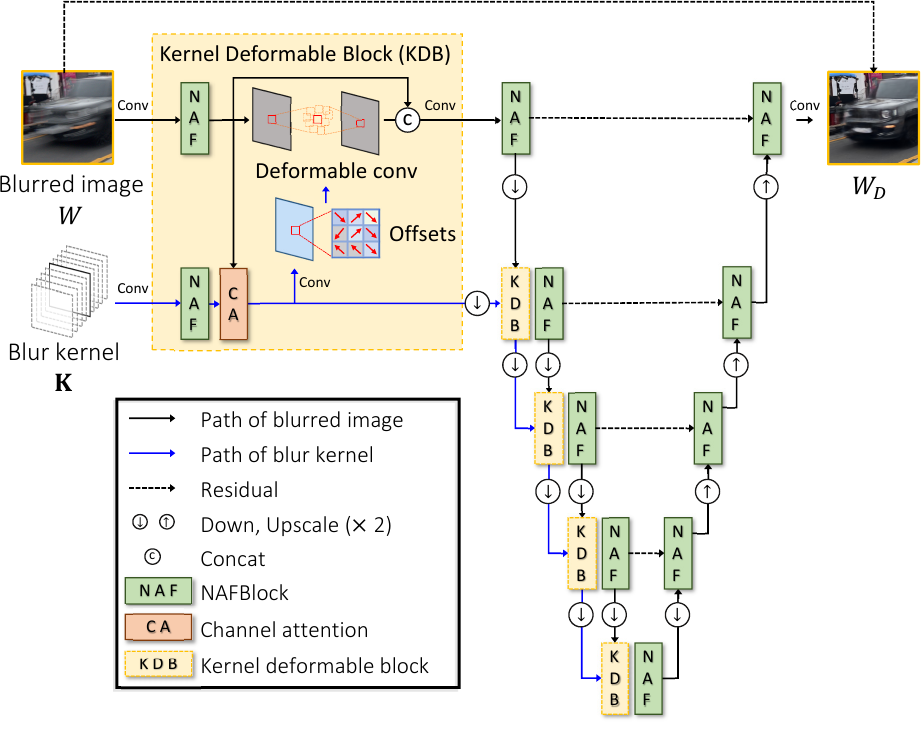}
    \caption{Architecture of HC-DNet.
    }
    \label{fig:HC-Dnet}
\end{figure}

\paragraph{HC-DNet}
HC-DNet takes a wide image $W$ and pixel-wise blur kernels $\mathbf{K}$ as inputs and produces a deblurred wide image $W_{D}$ (\Fig{HC-Dnet}).
The network adopts a U-Net architecture, in which we compose each level of the encoder and decoder networks with NAFBlocks~\cite{chen2022simple}.
To exploit $\mathbf{K}$, we design a novel kernel deformable block (KDB), which we attach before each level of the encoder.
In KDB, the features of $W$ and $\mathbf{K}$ are initially extracted by NAFBlocks. 
Channel attentions~\cite{hu2018senet} are then computed from the features of $W$ and $\mathbf{K}$ that are attended on features of blur kernels, which is to prevent utilization of potentially inaccurate blur kernels.
Subsequently, the attended features of $\mathbf{K}$ are used for predicting offsets and weights of a deformable convolution layer~\cite{zhu2019deformable}, which adaptively handles the features of $W$ according to $\mathbf{K}$, leading to more effective utilization of $\mathbf{K}$ in the deblurring process.
Then, the feature of $W$ and the output from the deformable convolution layer are concatenated and subsequently processed through a convolutional layer.

\subsection{Burst Image-based Refinement using HC-FNet}
\label{sec:hc-fnet}

\begin{figure}[!t]
    \centering
    \includegraphics[width=0.94\linewidth]{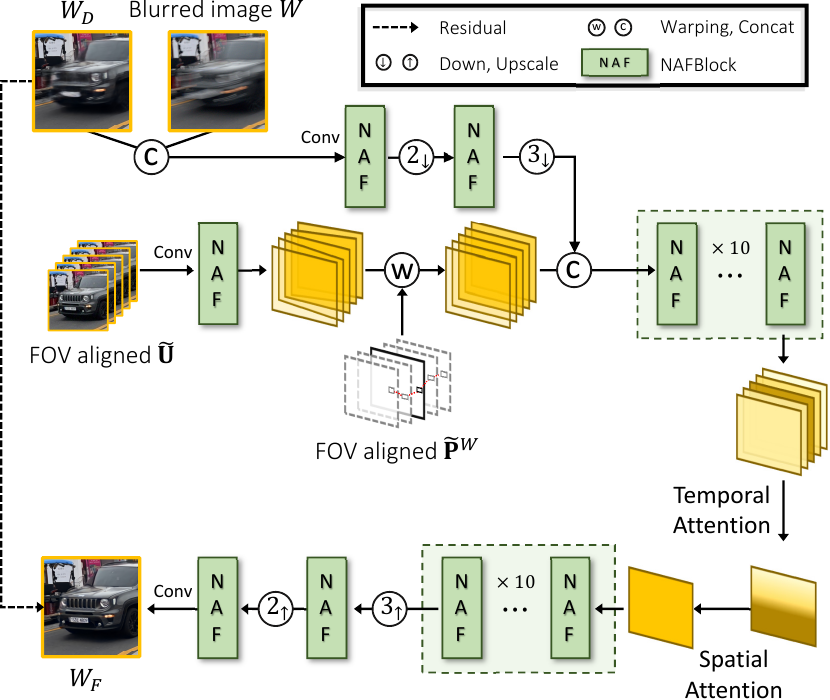}
    \caption{Architecture of HC-FNet.}
    \label{fig:HC-Fnet}
\end{figure}

\paragraph{Alignment of $\mathbf{U}$ to $W_{D}$}
For the fusion of $\mathbf{U}$ and $W_D$, we align $\mathbf{U}$ to the $6\times$ downsampled version of $W_D$, which is denoted by $W_{D,6\downarrow}$.
To this end, we warp both $\mathbf{U}$ and $\mathbf{P}$ using a $6\times$ downscaled version of $\hat{H}$, which is denoted as $\hat{H}_{6\downarrow}$,
and obtain FOV-aligned ultra-wide images $\tilde{\mathbf{U}} = \{\tilde{U}_1, \cdots, \tilde{U}_N\}$ and motion trajectories $\tilde{\mathbf{P}} = \{\tilde{P}_1, \cdots, \tilde{P}_N\}$.
Note that $\tilde{\mathbf{U}}$ are not downsampled but only aligned to match the FOV of the wide image.
Then, for more accurate alignment, we estimate the optical flow $\tilde{F}$ between $W_{D,6\downarrow}$ and the center frame $\tilde{U}_c$ of $\tilde{\mathbf{U}}$.
Finally, we compute the optical flow $\tilde{P}_i^W$ from $W_{D,6\downarrow}$ to the $i$-th FOV aligned ultra-wide image $\tilde{U}_i$ by combining $\tilde{P}_i$ and $\tilde{F}$.

\paragraph{HC-FNet} 
\Fig{HC-Fnet} shows the overall architecture of the HC-FNet, which is based on the architecture of a burst enhancement network~\cite{DBSR2021deep} and NAFBlock~\cite{chen2022simple}.
HC-FNet takes $\tilde{\mathbf{U}}$, and the deblurred wide image $W_{D}$ as input, and produces a refined wide image $W_F$.
HC-FNet also takes $W$ as additional input to obtain information that might be missing from $W_D$.
Then, HC-FNet extracts features from each $\tilde{U}_i$ using a NAFBlock and aligns the feature map to $W_D$ using $\tilde{P}^W_i$.
After the alignment, the features of $W_{D}$ and $W$ are concatenated to the feature map of each FOV-aligned ultra-wide image $\tilde{U}_i$.
These combined features of each ultra-wide image are then processed through NAFBlocks in a frame-wise manner.
We then merge the resulting features using the temporal and spatial attention module (TSA)~\cite{wang2019edvr} into a single frame.
Finally, the merged features are further processed and upsampled by additional NAFBlocks and upsampling layers.

\begin{table*}[t!]
\centering
\caption{Quantitative comparisons on the HCBlur-Syn and HCBlur-Real datasets. For the HCBlur-Real dataset, we use no-reference metrics (\ie, NIQE~\cite{mittal2012making}, BRISQUE~\cite{mittal2012no}, and TOPIQ~\cite{chen2023topiq} trained on SPAQ~\cite{SPAQ2020cvpr}) for evaluation. 
The number of parameters, MACs, and inference times of the kernel-based deblurring methods and HCDeblur include those for the optical flow estimation when the burst size of $\mathbf{U}$ is eight. 
HCDeblur$\bm{\mathit{_{small}}}$ indicates a lightweight version of HCDeblur, which uses a smaller version of RAFT~\cite{teed2020raft}.
}
\scalebox{1.0}{
\begin{tabular}{lcccccc}
\toprule[1.1pt]
                  & HCBlur-Syn             & HCBlur-Real                                     &             \\ \cmidrule(lr){2-2} \cmidrule(lr){3-3}
                  & \small{\;\; PSNR $\uparrow$ / SSIM $\uparrow$ \;\;} & \small{\;\; NIQE $\downarrow$ / BRISQUE $\downarrow$ / TOPIQ $\uparrow$ \;\;} & \small{Params. (M)}  & \small{MACs (G)}  & \small{Time (Sec.)} \\  \midrule[0.9pt]            
MIMO-UNet+~\cite{cho2021rethinking}        & 22.42 / 0.5966         & 5.49 / 27.87 / 0.44                             & 16.11 & 19563 & 4.52       \\
Uformer-B~\cite{Wang_2022_CVPR}         & 23.31 / 0.6332         & 5.17 / 29.63 / 0.55                             & 50.88 & 12483 & 15.11       \\
HINet~\cite{Chen_2021_CVPR}             & 23.60 / 0.6360         & 5.42 / 29.74 / 0.47                             & 88.67 & 21606 & 3.28       \\
NAFNet-32~\cite{chen2022simple}         & 23.61 / 0.6364         & 5.05 / 29.50 / 0.53                             & 17.11 & 2038 & 1.32       \\
NAFNet-64~\cite{chen2022simple}         & 24.18 / 0.6607         & 4.83 / 28.09 / 0.55                             & 67.89 & 8018 & 3.18       \\ \midrule
LSFNet~\cite{chang2021low}            & 22.32 / 0.6058         & 4.87 / 30.80 / 0.53                             & 10.20 & 4844 & 1.36       \\
D2HNet~\cite{zhao2022d2hnet}            & 22.64 / 0.6125         & 4.83 / 31.56 / 0.56                             & 79.58 & 9871 & 2.51       \\
NAFNet-Ref   & 24.37 / 0.6585         & 4.56 / 29.81 / 0.60                             & 18.14 & 2316 & 1.73       \\ \midrule
UFPNet~\cite{Fang_2023_CVPR}            & 24.44 / 0.6701         & 4.79 / 28.14 / 0.56                             & 78.37 & 20105 & 6.30       \\
MotionETR~\cite{zhang2021exposure}          & 25.24 / 0.6951         & 4.52 / 26.00 / 0.63                             & 10.68 & 11439 & 4.68       \\ \midrule
HCDeblur$_{small}$ (only HC-DNet)           & 25.80 / 0.7144         & 4.14 / 25.66 / 0.67                             & 10.62 & 2186 & 1.71       \\ 
HCDeblur$_{small}$  & 26.50 / 0.7281         & 3.97 / 24.79 / 0.69                             & 12.69 & 3286 & 3.16       \\
HCDeblur (only HC-DNet)           & 26.13 / 0.7251         & 4.11 / 25.51 / 0.67                             & 14.89 & 5189 & 2.35       \\ 
HCDeblur & 26.76 / 0.7373         & 3.95 / 24.65 / 0.69                             & 16.95 & 6288 & 3.85       \\ \bottomrule[1.1pt]
\end{tabular}
}
\label{tab:main_table_syn_real}
\end{table*}

\section{HCBlur Dataset}

In this section, we introduce HCBlur-Syn and HCBlur-Real datasets.
To construct the HCBlur-Syn dataset, we simultaneously capture wide and ultra-wide sharp videos and synthesize blurred wide images by averaging wide video frames.
We developed an iOS app to simultaneously capture wide and ultra-wide videos.
We used wide and ultra-wide cameras of an iPhone 13 Pro Max as our camera system. 
The ultra-wide camera captures videos of 720P resolution ($\text{720} \times \text{1280}$) at 60 FPS.
Simultaneously, the wide camera captures videos of 4K resolution ($\text{2160} \times \text{3840}$) at 30 FPS, which is the maximum frame rate supported by the multi-camera API~\cite{multicamAPI}.
Na\"{i}vely averaging wide video frames of 30 FPS may result in discontinuous blur due to significant motion between frames. 
To address this problem, we estimate optical flows between consecutive frames of wide videos and exclude the frames with a maximum displacement greater than 36 pixels.
Then, blurred wide images $W$ are generated by averaging the remaining frames of wide videos, and their temporal-center frames are chosen as ground-truth images. The number of averaged frames is randomly sampled from the set of $\{5, 7, 9, 11, 13\}$.

Then, for each blurred wide image $W$, ultra-wide images corresponding to $W$ are extracted from the ultra-wide videos using their timestamps.
Afterward, we randomly select a subset of these extracted burst images, ranging from 5 to 14 frames, to maintain a reasonable number of burst images.
To simulate noise on images, we synthesize Poisson-Gaussian noise on $W$ and $\mathbf{U}$ using shot and read noise parameters calibrated from our camera system.
Additionally, we adopt a realistic blur synthesis pipeline of Rim~et al.~\shortcite{rim_2022_ECCV} to improve the generalization ability of HCBlur-Syn to real-world images. This pipeline includes frame interpolation and saturated pixels synthesis. More details on synthesizing the HCBlur-Syn dataset can be found in the supplementary material.

Through the process described above, we collected 1,176 pairs of wide and ultra-wide videos of various indoor, daytime, and night-time scenes, and synthesized 8,568 pairs of a wide blurred image and a sequence of ultra-wide images for constructing the HCBlur-Syn dataset.
For training and evaluating deblurring methods, we split the HCBlur-Syn dataset into train, validation, and test sets, each of which consists of 5,795, 880, and 1,731 pairs, respectively.

The HCBlur-Real dataset provides 471 pairs of real-world $W$ and $\mathbf{U}$ from 157 of various night and indoor scenes. 
To collect the dataset, we simultaneously captured wide images and ultra-wide images using the iOS app for hybrid camera capturing (\Sec{camera_system}). 
Since ground-truth images are not included in the dataset, we utilize the HCBlur-Real dataset for evaluation using non-reference metrics~\cite{mittal2012making, mittal2012no, chen2023topiq}.

\section{Experiments}

\paragraph{Implementation Details} We utilize AdamW optimizer~\cite{kingma2014adam} and PSNR loss~\cite{chen2022simple} for training our method. The initial learning rate is set to 1e-3 and is gradually reduced to 1e-7 using the cosine annealing strategy~\cite{loshchilov2016cosine}.
We train our network in two steps; we first train HC-DNet and then train HC-FNet using the deblurred results of HC-DNet.
HC-DNet is trained for 300,000 iterations with a batch size of 32 and HC-FNet is trained for 150,000 iterations with a batch size of 8.
During the optimization of HC-FNet, the optical flow network~\cite{teed2020raft} for the alignment of $\mathbf{U}$ to $W_{D}$ is also jointly optimized.
The patch size for training is set to 384 for both HC-DNet and HC-FNet.

\subsection{Comparison with State-of-the-Art Methods}

To verify the effectiveness of HCDeblur, we evaluate HCDeblur and other deblurring methods on the HCBlur-Syn and HCBlur-Real datasets.
We compare HCDeblur with various deblurring methods including single-image, reference-based, and explicit blur kernel-based deblurring methods.
Single-image deblurring methods in our evaluation include CNN-based~\cite{cho2021rethinking,Chen_2021_CVPR,chen2022simple} and transformer-based~\cite{Wang_2022_CVPR} methods.
The transformer-based method~\cite{Wang_2022_CVPR} is evaluated using patch-wise testing due to the memory issue on 4K-resolution images.

Regarding reference-based deblurring methods, we include LSFNet \cite{chang2021low} and D2HNet~\cite{zhao2022d2hnet}, both of which exploit a short-exposure image for deblurring a long-exposure image.
As both models assume short-exposure images of the same spatial size and zoom ratio as those of the long-exposure image, we align and upscale the temporal center of ultra-wide images, and feed it to them as the short-exposure input.
Most single-image deblurring networks can be easily extended to take an additional reference image.
Thus, to cover such a case in our comparison, we also include a variant of NAFNet-32~\cite{chen2022simple} that is modified for reference-based deblurring, which is denoted by NAFNet-Ref.
Specifically, NAFNet-Ref aligns the temporal center of ultra-wide images to a blurred wide image using RAFT~\cite{teed2020raft}, and both wide and aligned ultra-wide images are fed to a NAFNet-32.
For evaluation, we train both RAFT and NAFNet-32 in an end-to-end manner.
More details of NAFNet-Ref are provided in the supplementary material.

Finally, as HC-DNet leverages blur kernels, we also include recent explicit blur kernel-based methods: MotionETR~\cite{zhang2021exposure} and UFPNet~\cite{Fang_2023_CVPR} in our evaluation.
Both methods estimate blur kernels using single-image blur kernel estimation networks and use the blur kernels in deblurring networks.
We modified them to use blur kernels estimated from ultra-wide images instead of their kernel estimation networks.
All the compared methods are trained on the HCBlur-Syn dataset for fair comparisons.

\Tbl{main_table_syn_real} shows a quantitative comparison of HCDeblur and the other methods on the HCBlur-Syn and HCBlur-Real datasets.
HCDeblur exhibits superior performance compared to all the other methods on both datasets.
Notably, HC-DNet alone still outperforms all the competitors.
Our final model achieves the best performance as HC-FNet further refines the deblurred results using the ultra-wide burst images. 
We also compare a smaller version of HCDeblur: HCDeblur$_{small}$, which uses a small version of RAFT~\cite{teed2020raft} for constructing motion trajectories.
As the table shows, HCDeblur$_{small}$ still outperforms all the other methods.
This result clearly proves the advantage of utilizing ultra-wide burst images.

Interestingly, the reference-based deblurring methods do not significantly outperform the single-image deblurring methods despite using additional reference images.
This is because our hybrid camera system captures low-resolution ultra-wide images while these methods require high-resolution reference images, and also because they are sensitive to the alignment errors between input and reference images.
A detailed analysis of the sensitivity of reference-based deblurring on the alignment error is provided in the supplementary material.
While NAFNet-Ref performs the best among the single-image and reference-based deblurring methods, it still performs worse than ours.

Thanks to the motion information in the ultra-wide burst images, the explicit blur kernel-based methods, UFPNet~\cite{Fang_2023_CVPR} and MotionETR~\cite{zhang2021exposure}, outperform the single-image and reference-based deblurring methods.
Nevertheless, HCDeblur still outperforms both UFPNet and MotionETR thanks to its more carefully designed network architecture of HC-DNet, and the refinement process by HC-FNet.

\begin{figure*}[!t]
    \centering
    \includegraphics[width=1.0\linewidth]{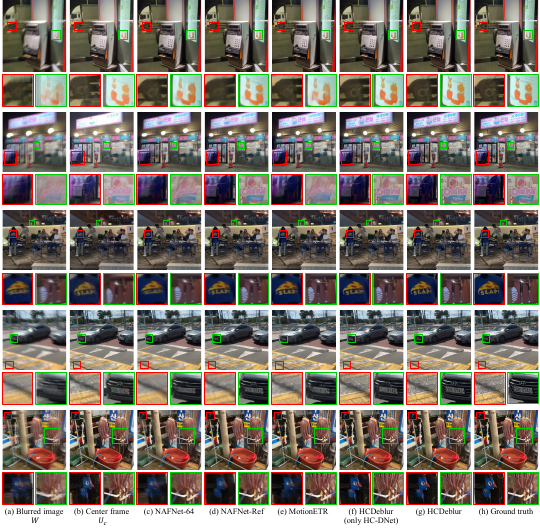}
    \setlength{\abovecaptionskip}{-2.1mm}
    \caption{Qualitative comparisons on the HCBlur-Syn dataset. We compare HCDeblur with NAFNet-64~\cite{chen2022simple}, NAFNet-Ref, and MotionETR~\cite{zhang2021exposure}.}
    \label{fig:main_qualitative}
\end{figure*}

\begin{figure*}[!t]
    \centering
    \includegraphics[width=1.0\linewidth]{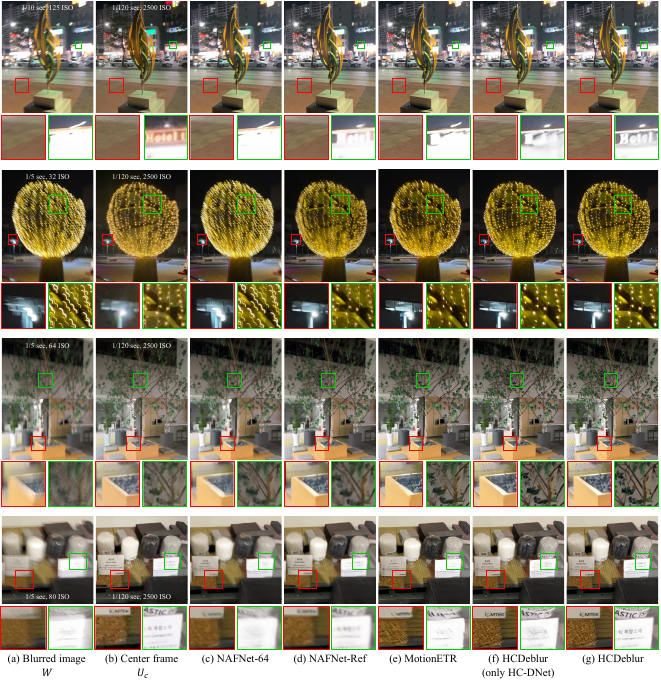}
    \setlength{\abovecaptionskip}{-2.1mm}
    \caption{Qualitative comparison on the HCBlur-Real dataset. We compare HCDeblur with NAFNet-64~\cite{chen2022simple}, NAFNet-Ref, and MotionETR~\cite{zhang2021exposure}.}
    \label{fig:real_qualitative}
\end{figure*}

\Figs{main_qualitative} and \ref{fig:real_qualitative} show qualitative evaluations on the HCBlur-Syn and HCBlur-Real datasets, respectively.
As the figures show, HCDeblur successfully deblurs 4K images, restoring sharp details of small texts and edges.
In contrast, NAFNet-64~\cite{chen2022simple} and NAFNet-Ref fail to restore sharp details. 
MotionETR~\cite{zhang2021exposure} sometimes fails to restore sharp details because of its sensitivity to inaccuracies in blur kernels. It uses the trajectories of blur kernels as fixed offsets in deformable convolution~\cite{zhu2019deformable}, making it vulnerable to inaccurate blur kernels. 
In contrast, HC-DNet effectively addresses errors in blur kernels by learning the deformable convolution offsets from blur kernels, resulting in much sharper deblurred images. Our final model further enhances deblurred images using HC-FNet and achieves the sharpest results.

\begin{table}[t]
    \centering
    \caption{
    Ablation study for variants of HCDeblur. We compare its performance without HC-DNet and with NAFNet-32~\cite{chen2022simple}, evaluate different alignment modules (RAFT~\cite{teed2020raft} vs BIPNet~\cite{dudhane2021burst}) after the FOV alignment, and examine burst features fusion strategies like averaging (Avg.), transposed attention (Trans Att.)~\cite{mehta2023gated}, and temporal and spatial attention (TSA)~\cite{wang2019edvr}.
    }
    \scalebox{0.90}{
        \begin{tabular}{cccc}
            \toprule[1.2pt]
            \small DeblurNet    & \small Alignment  & \small FusionNet           &\small PSNR $\uparrow$ / SSIM $\uparrow$\\ \midrule[1.0pt]
            $\times$  &  FOV + RAFT       & HC-FNet + TSA             & 23.22 / 0.5963 \\
            NAFNet-32 &  FOV + RAFT       & HC-FNet + TSA             & 24.05 / 0.6493 \\ \cmidrule(lr){1-4}
            HC-DNet   & FOV & HC-FNet + TSA              & 26.63 / 0.7369 \\ 
            HC-DNet   &  FOV + BIPNet & HC-FNet + TSA              & 26.66 / 0.7370 \\ \cmidrule(lr){1-4}
            HC-DNet   &   FOV + RAFT       & HC-FNet + Avg.              & 26.66 / 0.7379 \\ 
            HC-DNet   &   FOV + RAFT       & HC-FNet + Trans Att.   & 26.70 / 0.7367 \\ 
            HC-DNet   & FOV + RAFT       & HC-FNet + TSA              & 26.76 / 0.7373 \\  \bottomrule[1.2pt]
        \end{tabular}
    }
    \label{tab:ablation_hcfnet}
\end{table}

\subsection{Ablation Studies}
\label{sec:ablation}

\Tbl{ablation_hcfnet} shows the effects of different components of HCDeblur.
To evaluate the effect of HC-DNet, we also prepare two variants of HCDeblur: one without HC-DNet and the other with NAFNet-32~\cite{chen2022simple}, which uses an input blurred image $W$ and the deblurred result of NAFNet-32 instead of $W_D$, respectively.
Both methods significantly underperform the original HCDeblur model, validating the importance of HC-DNet. We also compare different alignment strategies for HC-FNet in \Tbl{ablation_hcfnet}.
In the table, FOV means FOV alignment, and BIPNet indicates the alignment module proposed by Dudhane~et al.~\shortcite{dudhane2021burst} that implicitly aligns features using deformable convolution.
Again, among all the alignment strategies, `FOV + RAFT' performs the best thanks to the high-quality optical flow maps estimated from a deblurred output $W_D$ and $\mathbf{U}$.
Finally, we also examine different feature fusion strategies for HC-FNet.
As the table shows, temporal and spatial attention (TSA)~\cite{wang2019edvr} achieves better performance than the other fusion strategies.

\paragraph{Kernel Deformable Block} 
In HC-DNet, KDB is a crucial component for exploiting blur kernels.
Here, we verify the effectiveness of the component by comparing it with other approaches: simple concatenation, the kernel guided convolution (KGC)~\cite{Kaufman_2020_CVPR}, the kernel attention module (KAM)~\cite{Fang_2023_CVPR}, and the motion aware block (MAB)~\cite{zhang2021exposure}.
The simple concatenation indicates that features of blur kernels and a blurred image are concatenated and merged by a CNN.
KGC utilizes blur kernels to linearly transform features of a blurred image, while KAM employs blur kernels to determine the weights of pixel-wise channel attention.
MAB directly uses the trajectories of blur kernels as fixed offsets of deformable convolution~\cite{zhu2019deformable}. In contrast, KDB learns to estimate the offsets of deformable convolution kernels from the features of blur kernels.
We applied each approach to the encoder of HC-DNet instead of KDB.
\Tbl{hcdnet_ablation} shows KDB achieves the highest PSNR and SSIM scores, proving the effectiveness of KDB.

\begin{table}[t]
    \centering
    \caption{Comparison of different schemes for exploiting blur kernels.
    }
    \scalebox{1.0}{
    \begin{tabular}{lc}
        \toprule[1.2pt]
        Methods & PSNR $\uparrow$ / SSIM $\uparrow$ \\ \midrule[1.0pt]
        HC-DNet                  & 23.18 / 0.6184 \\
        + KAM~\cite{Fang_2023_CVPR}      & 25.35 / 0.6909 \\
        + Concat          & 25.39 / 0.6920 \\
        + KGC~\cite{Kaufman_2020_CVPR}      & 25.45 / 0.6953 \\
        + MAB~\cite{zhang2021exposure} & 25.52 / 0.7004 \\
        + KDB w/o Channel Attention      & 26.05 / 0.7229 \\
        + KDB & 26.13 / 0.7251 \\ \bottomrule[1.2pt]
    \end{tabular}
    }
    \label{tab:hcdnet_ablation}
\end{table}

\section{Conclusion}
In this paper, we proposed HCDeblur, a novel learning-based hybrid camera deblurring framework for smartphone cameras.
Our framework adopts HC-DNet and HC-FNet, which utilize burst ultra-wide images for blur kernel estimation and refinement of a deblurred image, respectively.
Additionally, we have also presented the HCBlur dataset for training and evaluation of the proposed method.
Our experiments demonstrate the effectiveness of the proposed method.

\paragraph{Limitations} Our framework uses a large optical flow network, RAFT~\cite{teed2020raft}, and computes optical flow across burst images, increasing the computational overhead. The inference time of our system is about four sec., which is too slow to serve as an on-device application on smartphones. However, the optical flow network can be replaced by any efficient network, and HCDeblur could be further optimized using acceleration techniques.
Another limitation is that, as our method relies on optical flow, its performance may degrade when optical flow estimation fails, e.g., it may fail for extremely noisy or textureless images.
To resolve these issues would be interesting future directions.

\begin{acks}
This work was supported by Samsung Research Funding \& Incubation Center
of Samsung Electronics under Project Number SRFC-IT1801-52 and the National Research Foundation of Korea (NRF) grants (No.2023R1A2C200494611) funded by the Korea government (MSIT) and Institute of Information \& communications Technology Planning \& Evaluation (IITP) grants (No.2019-0-01906, Artificial Intelligence Graduate School Program (POSTECH)) funded by the Korea government (MSIT).
\end{acks}

\bibliographystyle{ACM-Reference-Format}
\bibliography{sample-bibliography}

\clearpage

\clearpage

\end{document}